\documentclass[10pt,twocolumn,letterpaper]{article}

\usepackage{iccv}
\usepackage{times}
\usepackage{epsfig}
\usepackage{amsmath}
\usepackage{amssymb}
\usepackage{array,multirow,graphicx}
\usepackage{color}
\usepackage{caption}
\usepackage{subcaption}
\usepackage{MnSymbol}

\usepackage[pagebackref=true,breaklinks=true,letterpaper=true,colorlinks,bookmarks=false]{hyperref}

\iccvfinalcopy 

\def\iccvPaperID{5793} 

\ificcvfinal\fi

\newcolumntype{?}{!{\vrule width 1pt}}

\begin{document}

\title{SMPLR: Deep SMPL reverse for 3D human pose and shape recovery}

\author{\parbox{16cm}{\centering
    {\large Meysam Madadi$^{1}$, Hugo Bertiche$^{2}$ and Sergio Escalera$^{1,2}$}\\
    {\normalsize
    $^1$ Computer Vision Center, Edifici O, Campus UAB, 08193 Bellaterra (Barcelona), Catalonia Spain\\
    $^2$ Dept. Mathematics and Informatics, Universitat de Barcelona, Catalonia, Spain}}
}

\maketitle

\begin{abstract}
Current state-of-the-art in 3D human pose and shape recovery relies on deep neural networks and statistical morphable body models, such as the Skinned Multi-Person Linear model (SMPL). However, regardless of the advantages of having both body pose and shape, SMPL-based solutions have shown difficulties to predict 3D bodies accurately. This is mainly due to the unconstrained nature of SMPL, which may generate unrealistic body meshes. Because of this, regression of SMPL parameters is a difficult task, often addressed with complex regularization terms. In this paper we propose to embed SMPL within a deep model to accurately estimate 3D pose and shape from a still RGB image. We use CNN-based 3D joint predictions as an intermediate representation to regress SMPL pose and shape parameters. Later, 3D joints are reconstructed again in the SMPL output. This module can be seen as an autoencoder where the encoder is a deep neural network and the decoder is SMPL model. We refer to this as SMPL reverse (SMPLR). 
By implementing SMPLR as an encoder-decoder we avoid the need of complex constraints on pose and shape. Furthermore, given that in-the-wild datasets usually lack accurate 3D annotations, it is desirable to lift 2D joints to 3D without pairing 3D annotations with RGB images. Therefore, we also propose a denoising autoencoder (DAE) module between CNN and SMPLR, able to lift 2D joints to 3D and partially recover from structured error. We evaluate our method on SURREAL and Human3.6M datasets, showing improvement over SMPL-based state-of-the-art alternatives by about 4 and 25 millimeters, respectively.
\end{abstract}

\vspace{-0.5cm}
\section{Introduction}
\vspace{-0.1cm}
\label{sec:introduction}

3D human pose estimation from still RGB images is a challenging task due to changes in lighting conditions, cluttered background, occlusions, inter and intra subject pose variability, as well as ill-posed depth ambiguity. Moreover, due to its nature, accurate annotation of captured data is not a trivial task and most available datasets are captured under controlled environments \cite{h36m_2014, sigal2010humaneva,mehta2017vnect}. 


\begin{figure}[!t]
    \centering
    \includegraphics[width=\linewidth]{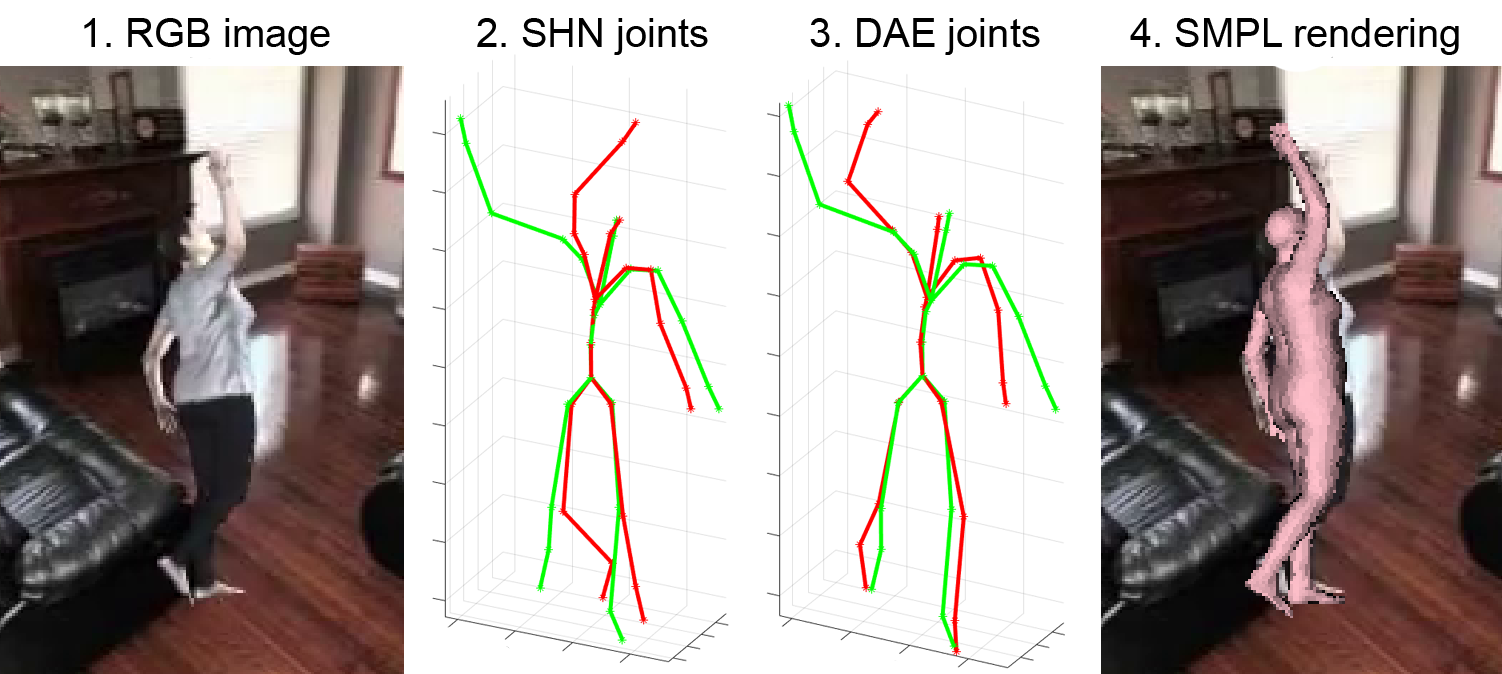}
    \vspace{-0.7cm}\caption{Illustration of the proposed model output. Given 1) an input RGB image, 2) an initial estimation of 3D joint locations is applied based on a given CNN (volumetric stacked hourglass (SHN) \cite{newell2016stacked} in this case). Green line shows ground truth and red line shows 3D estimated joints. 3) Our denoising autoencoder model is able to recover pose from structured error. Finally, 4) body mesh is rendered by SMPL based on the proposed SMPL reverse strategy.}    \vspace{-0.3cm}
    \label{fig:motivation}
\end{figure}

One case of human pose representation is 3D joint locations. 
However, 3D joints do not implicitly show the morphology of the body. Being able to estimate body shape or mesh along with joints allows a wide method applicability, including movie editing, body soft-biometrics measurements or cloth retexturing, among others. Besides, such a dense body representation may help to achieve more accurate estimations of 3D joints. 
Available body models range from simple geometrical objects, like compositions of cylinders and spheres, to complex parametric statistical models such as SCAPE~\cite{anguelov2005scape} and SMPL~\cite{loper2015smpl}. SMPL~\cite{bogo2016keep,lassner2017unite,pavlakos2018learning,kanazawa2018end,omran2018neural} generates realistic body meshes based on PCA shape components along with relative axis-angle rotations of joints. Rotations form body pose in a defined kinematic tree and are computed for joints with respect to their parent's nodes. 
The goal is to estimate SMPL parameters from an RGB image such that generated body mesh describes and fits as much as possible to the visible human in the image. This can be done by fitting generative models ~\cite{sigal2008combined,ivekovivc2008human,guan2009estimating,bogo2016keep,lassner2017unite} or training discriminative deep models~\cite{pavlakos2018learning,kanazawa2018end,omran2018neural,varol18_bodynet}. On the one hand, regular generative optimization solutions are shown to be sensitive to noise and need a careful and complex design of objective functions. On the other hand, while deep models have shown superior performance over the former solutions, they are data-hungry approaches. In both cases, a direct regression of SMPL parameters is a complex task because: 1) SMPL is a many-to-one complex function\footnote{SMPL details are given in the section \ref{sec:smpl} } which is sensitive to data noise (i.e. optimizing SMPL parameters may converge to invalid values), and 2) accurate image annotation with SMPL pose and shape parameters in large in-the-wild datasets is infeasible. Therefore, researchers developed their solutions based on available 2D joints by applying intermediate representations~\cite{pavlakos2018learning,omran2018neural} or adversarial training~\cite{kanazawa2018end} for 3D inference. However, it is known that estimation of 3D data from 2D is ill-posed and can lead to sub-optimal solutions.

In this paper, given an RGB image, we first estimate 3D joints and a sparse set of landmarks placed along body surface. Then, we use them to regress SMPL pose and shape parameters. The output is a detailed body mesh. We call this procedure SMPL reverse (SMPLR). One can imagine SMPLR as an autoencoder where latent embeddings are pose and shape components. We define encoder as a number of Multi-layer Perceptron (MLP) networks while decoder is SMPL. By first estimating 3D joints and landmarks, as an intermediate representation, 1) we avoid a direct regression of SMPL parameters, which easily yields non-realistic body meshes, 2) we can safely train SMPLR, even end-to-end, in a simple way without explicit constraints on SMPL, and 3) we provide flexibility to the design, i.e. SMPLR can be trained independently using millions of generated mocap-like data, thus allowing cross-dataset generalization.

When 3D ground truth data is available for RGB images, any state-of-the-art CNN can be used to estimate 3D joints and landmarks. However, such ground truth data is not available for in-the-wild datasets. Besides, estimated 3D joints can have structured error due to depth ambiguity or occlusions. To handle these cases we design a denoising autoencoder (DAE) network \cite{vincent2010stacked} as an extra module between CNN and SMPLR able to lift 2D joints to 3D and/or recover from structured error. We show the proposed model output in Fig. \ref{fig:motivation}. 

In summary, our main contributions are as follows:
\begin{itemize}\vspace{-0.3cm}
    \item We build a denoising autoencoder that learns to recover input data from structured error. The model transforms 2D/3D joints to a more human-consistent 3D joint predictions, enforcing symmetry and proportions on bone lengths.
    \vspace{-0.3cm}
    \item We design a two-branch MLP network to regress SMPL parameters from 3D joints and landmarks given by DAE. We refer to the combination of DAE+MLP+SMPL as SMPLR. This allows the inference of human body mesh from a sparse point representation. Finally, we gain an improvement over chosen CNN by end-to-end training with SMPLR. 
    \vspace{-0.3cm}
    \item Throughout our experiments, we demonstrate that it is possible to obtain an accurate human body model from a set of joint and landmark predictions. We obtain state-of-the-art results for SMPL-like architectures on Human3.6M \cite{h36m_2014} and SURREAL \cite{varol17_surreal} datasets.
\end{itemize}


\vspace{-0.1cm}
\section{Related work}
\label{sec:related}
\vspace{-0.1cm}
In this section, we review state-of-the-art works on 3D human pose estimation from still RGB images.

\textbf{Lifting 2D to 3D. } Depth regression from 2D 
is an ill-posed problem where several 3D poses can be projected to the same 2D joints. Chen and Ramanan \cite{chen20173d} show that copying depth from 3D mocap data can provide a fair estimation when a nearest 2D matching is given. However, Moreno \cite{moreno20173d} shows that distance of random pairs of poses has more ambiguity in Cartesian space than Euclidean distance matrix. 
Recent works show that directly using simple \cite{martinez2017baseline} or cascade \cite{hoang2018cascade} MLP networks can be more accurate. Additionally, 2D joints can be wrongly estimated, making previous solutions sub-optimal. Yang \etal \cite{yang20183d} use adversarial training and benefit from available 3D data along with 2D data to infer depth information. 
In our case, the proposed denoising autoencoder is used to lift 2D pose to 3D in the lack of paired image and 3D ground truth data.

\textbf{Direct regression} refers to regressing 3D pose directly from an RGB image. Due to the nonlinear nature of the human pose, 3D pose regression without modeling correlation of joints is not a trivial task. 
Brau and Jiang \cite{brau20163d} estimate 3D joints and camera parameters without direct supervision on them. Instead, they use several loss functions for projected 2D joints, bone sizes and independent Fisher priors. Sun \etal \cite{sun2017compositional} propose a compositional loss function based on relative joints with respect to a defined kinematic tree. They separate 2D joints and depth estimation in the loss. We avoid relying on such complex losses by using 3D joints and landmarks as intermediate representation. 

\begin{figure*}[!ht]
    \centering
    \includegraphics[width=\linewidth]{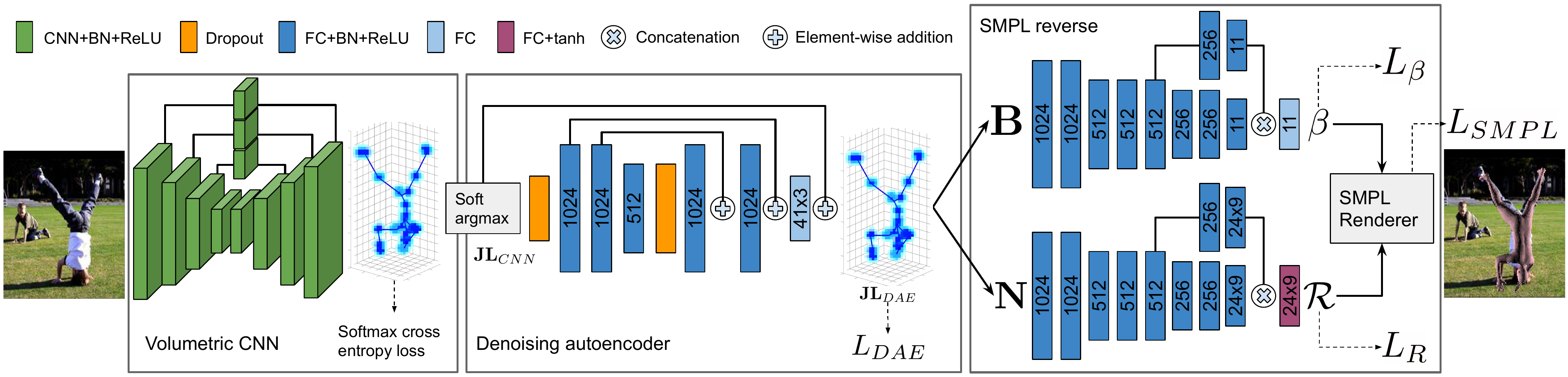}
\vspace{-0.7cm}    \caption{System pipeline. A CNN estimates volumetric heatmaps. Soft argmax converts heatmaps to joints locations and feeds them to denoising autoencoder module. Soft argmax is differentiable, thus, gradients can be backpropagated. Finally, we compute normalized relative distances $\mathbf{B}$ (eq. \ref{eq:bone}) and normalized relative joints $\mathbf{N}$ (eq. \ref{eq:N}) which are fed to two independent networks designed to regress SMPL parameters. At the end SMPL is responsible to render a realistic body mesh. $\dashedrightarrow$ shows where the loss is applied.}
    \label{fig:pipeline}\vspace{-0.3cm}
\end{figure*}

\textbf{Probability maps} are joints likelihood computed for each pixel/volume. From 2D joint heatmaps different solutions are applied to infer the third dimension. Tome \etal \cite{tome2017lifting} iteratively minimize a function based on 2D belief map and a learnt 3D model to find the most likely 3D pose. 
Since probability maps are dense predictions, fully convolutional networks are usually applied. Luo \etal \cite{luo2018orinet} extend stacked hourglass (SHN) \cite{newell2016stacked} to estimate 2D joint heatmaps along with limb orientation map in each stack. Pavlakos \etal \cite{pavlakos2017coarse} extend SHN to output 3D volumetric data by a coarse-to-fine architecture where at each stack the third dimension is linearly increased with 2D heatmaps. 
Nibali \etal \cite{nibali20183d} propose marginal heatmaps from different axis viewpoints. 

\textbf{Pose and shape estimation. } In order to compute a detailed body model, it is common to estimate body volume or mesh along with 3D pose. Early proposals were mainly based on the combination of simple geometric objects \cite{stoll2011fast,sigal2012loose}, while recent approaches are PCA-based parametric models like SMPL \cite{loper2015smpl}. Bogo \etal \cite{bogo2016keep} were the first to apply SMPL for body pose and shape recovery. Their method was based on regular optimization procedures given an objective function with several constraints, minimizing projected joints and pre-estimated 2D joints. 
Such a complex function design is critical for the success of optimization procedures, since SMPL is a many-to-one function and sensitive to noise. Lassner \etal \cite{lassner2017unite} extended the previous work by including a bi-directional distance between projected mesh and body silhouette. Recent works embed SMPL within deep models. Tung \etal \cite{tung2017self} regressed SMPL pose and shape along with camera parameters. They trained the model with supervision on synthetic data and fine-tuned without supervision at inference time using 2D joints, silhouette and motion losses. Similarly, Kanazawa \etal \cite{kanazawa2018end} regressed as well SMPL and camera parameters, but in addition, they applied adversarial training. Predictions are fed to a discriminator network which classifies them as real/fake with respect to real body scans. 
Similar to our work, \cite{pavlakos2018learning,omran2018neural,zanfir2018deep} estimate pose and shape parameters from intermediate information, like body segments \cite{omran2018neural}, 2D joint heatmaps and body mask \cite{pavlakos2018learning} or 3D joints \cite{zanfir2018deep}. They include SMPL to obtain 3D joints and mesh which are used to compute the loss either in 2D (back-projected from 3D) or 3D. However, this process is ill-posed and sub-optimal because of the loss of depth information (in the case of \cite{pavlakos2018learning,omran2018neural}) or the ambiguity in joint orientations and shape parameters (in the case of \cite{zanfir2018deep}).
We show 3D joints and surface landmarks can better deal with this problem outperforming aforementioned solutions. Recently, Varol \etal \cite{varol18_bodynet} proposed a multi-tasking approach to estimate body 2D/3D pose, pixel segments and volumetric shape. They use SMPL to generate ground truth body volumes and do not embed SMPL function within the network. 

In this paper, we propose an approach to estimate 3D body pose and shape by the use of intermediate information and SMPL model. We can benefit from end-to-end training. Besides, our method can be adapted to 2D-to-3D solutions when just 2D ground truth data is available.

\vspace{-0.1cm}\section{Methodology}\vspace{-0.1cm}
\label{sec:methodology}
We estimate 3D joints $\mathbf{J}=\{j\}_1^K$ and surface vertices $\mathbf{T}=\{t\}_1^C$ from a single RGB image $\mathit{I}$, where $j,t\in\mathbb{R}^3$, and $K$ and $C$ are the number of body joints and surface points, respectively. We define $\mathbf{L}\subset\mathbf{T}$ as a set of sparse surface landmarks and $\mathbf{JL}$ as a concatenation of the two matrices. In order to compute a detailed mesh, we use SMPL model \cite{loper2015smpl}. Our goal is to estimate SMPL parameters from image $\mathit{I}$ using deep learning without directly regressing them. This way, we avoid possible artifacts while keeping the architecture flexible in the lack or presence of 3D ground truth data. Our network contains three main modules shown in Fig. \ref{fig:pipeline}. First, joint and landmark locations are estimated by any chosen CNN ($\mathbf{JL}_{CNN}$). Afterwards, DAE filters structured error or lifts 2D to 3D ($\mathbf{JL}_{DAE}$). Finally, SMPLR recovers pose and shape from the predictions of the previous module and reconstructs a detailed body mesh and 3D joints. Next, we explain DAE and SMPLR in detail.

\vspace{-0.1cm}\subsection{SMPL review}\label{sec:smpl}\vspace{-0.1cm}
SMPL is a statistical parametric function $\mathit{M}(\beta,\theta;\Phi)$ which maps shape parameters $\beta$ and axis-angle pose parameters $\theta$ into vertices $\mathbf{T}$, given learnt model parameters $\Phi$. Given a template average body mesh with vertices $\mathbf{T}^*$ and a dataset of scanned bodies, two sets of principal components $\mathcal{S}=[\mathbf{S}_1,...,\mathbf{S}_{|\beta|} ]\in \mathbb{R}^{3C\times|\beta|}$ and $\mathcal{P}=[\mathbf{P}_1,...,\mathbf{P}_{9K} ]\in \mathbb{R}^{3C\times9K}$ are learnt to form model parameters $\Phi$ (where $|\beta|=10$, $C=6890$ and $K=24$). Then, template shape vertices $\mathbf{T}^*$ can be morphed to $\mathbf{T}^{*}_{s}$ by ${vec}^{-1}_{3, C}(\mathcal{S}\times\beta)+\mathbf{T}^{*}$ where ${vec}^{-1}(.)$ is a reshaping operator. Bases $\mathbf{P}_i$ are responsible for small pose-based displacements due to body soft-tissue behavior and have small contribution in the shape deformation. Given a kinematic tree (i.e. Fig. \ref{fig:landmarks}) a set of relative rotation matrices $\mathcal{R}=[\mathbf{R}_1,...,\mathbf{R}_{K} ]\in \mathbb{R}^{3\times3}$ are computed for each joint with respect to their parents. Each $\mathbf{R}_i$ is a function of $\theta_i\in \mathbb{R}^3$ and is computed based on Rodrigues formulation. These rotation matrices are mainly used because of two reasons: i) to pose the mesh by rotating body parts relatively in the kinematic tree, and ii) to update the template shape in rest pose $\theta^{*}$ by basis $\mathbf{P}_i$. Rotations are applied based on joints. Nevertheless, joints are a function of body shape, therefore they need to be computed before posing the model. This is done by a regressor matrix $\mathcal{J}$ (as part of parameters $\Phi$) from updated vertices $\mathbf{T}^{*}_{s}$. Please read \cite{loper2015smpl} for more detailed explanations of the SMPL.

SMPL model has several characteristics. First, it is differentiable, which yields the possibility to be used along with deep networks. Secondly, it does not constrain invalid pose and shape values and, thus, it is a many-to-one function. This means that given an RGB image, end-to-end training of a CNN from scratch with SMPL attached on top may converge to a non-optimal solution. One of the main reasons of this is the usage of Rodrigues formulation and axis-angle pose parameters, as it is known not to be unique (periodicity of $\theta$). A possible solution is to directly use rotation matrices as proposed in \cite{lassner2017unite,omran2018neural}. Finally, SMPL is a generative model which allows us to generate mocap-like data for free. Also, it can be used to generate synthetic realistic images \cite{varol17_surreal}. 

\vspace{-0.1cm}\subsection{SMPL reverse}\vspace{-0.1cm}
A natural way of embedding SMPL in a deep network is to estimate $\beta$ and $\theta$ given image $\mathit{I}$ and feed them to SMPL. However, this is a challenging task because of the aforementioned many-to-one property, plus the noise sensitivity of the model. Besides, direct regression of SMPL parameters may generate artifacts \cite{kanazawa2018end,tan2017indirect}. Instead, researchers use intermediate representations like 2D joints and body silhouette \cite{pavlakos2018learning} or body segments \cite{omran2018neural} to regress SMPL parameters. Although such data is easier to annotate from RGB images than SMPL data, they provide sub-optimal mapping to SMPL parameters, because 1) estimating 3D from 2D is an ill-posed problem, and 2) the loss is computed from noisy back-projected 2D estimations. In this paper, we instead propose an autoencoder-like scheme, i.e. the input 3D data is recovered in the output, while pose and shape are obtained in the encoder and SMPL is taken as decoder. We refer to this model as SMPL reverse (SMPLR, see Fig. \ref{fig:pipeline}). 

This design has several benefits: SMPLR can be trained 1) without the need of constraints on SMPL, 2) independent to RGB data using millions of generated 3D mocap-like data, and 3) end-to-end with a CNN. All of these provide simplicity and flexibility in the design and training of the entire network. In the results section we show that SMPLR formulation can generate more accurate estimations than state-of-the-art SMPL-based alternatives. Furthermore, SMPLR acts as a strong regularization on CNN model when trained end-to-end and it enhances the internal coherence among joints for CNN predictions. In sec. \ref{sec:end-to-end ablation} we propose an effective incremental training for this task.

We model SMPLR encoder with deep MLP networks. We design two independent networks $\mathcal{R}=\Omega(\mathbf{N};\phi_p)$ and $\beta=\Psi(\mathbf{B};\phi_s)$ with the same architecture (see Fig. \ref{fig:pipeline} for details) for pose and shape estimation, respectively, where $\phi_.$ corresponds to network parameters, $\mathbf{N}$ is a vector of normalized relative joints and $\mathbf{B}$ is a vector of normalized relative distances. A reason for the choice of two networks is that $\mathcal{R}$ and $\beta$ are independent variables. Besides, we want the encoder to be cross-dataset applicable and explainable w.r.t. the pose and shape parameters. In available datasets, while pose parameters have a high variability, there is no much variability in shape parameters. For instance, Human3.6M dataset \cite{h36m_2014} only consists of 11 subjects and training $\Psi$ is not feasible by relying just on this dataset. In the results section we show the contribution of each network to the final joint estimates.

\begin{figure}[!t]
    \centering
    \begin{subfigure}[b]{0.4\textwidth}
        \includegraphics[width=\textwidth]{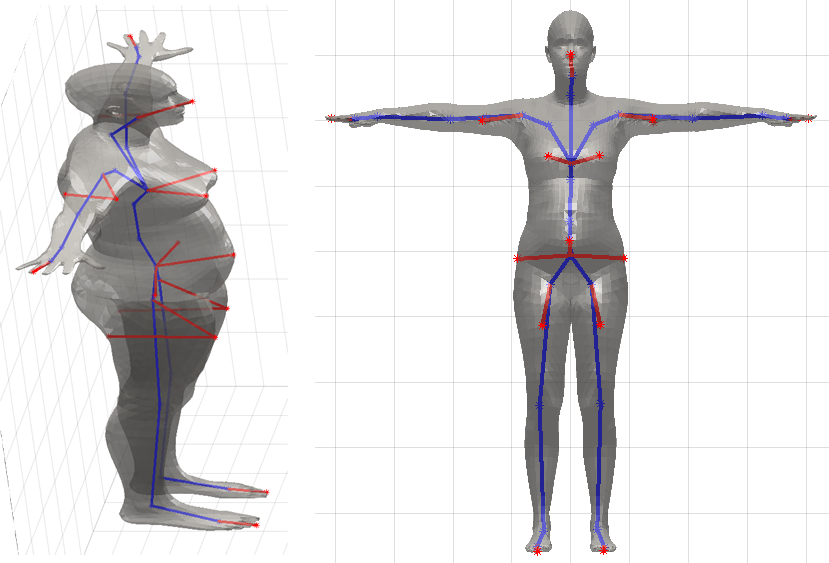}
    \end{subfigure}\vspace{-0.3cm}
    \caption{Landmarks and assigned kinematic tree in red (blue is original SMPL kinematic tree). Pelvis is set as root.}
    \label{fig:landmarks}\vspace{-0.4cm}
\end{figure}

Since we define SMPLR as an autoencoder, its input must be $\mathbf{J}$ and $\mathbf{T}$. However, $\mathbf{T}$ is a high dimensional vector and all vertices do not necessarily contribute equally in the computation of pose and shape parameters, wasting network capacity if all of them are considered. To cope with this issue, we empirically select a subset of points as landmarks $\mathbf{L}\subset\mathbf{T}$, which represent body shape and complement $\mathbf{J}$. Without landmarks, network converges to the average body fatness and the problem still remains ill-posed due to the ambiguity of joints orientation. Landmarks help to cope with these two problems. Besides, it is cheaper to gather landmarks in mocap datasets rather than scanning the whole body. We show the 18 selected landmarks and their assigned kinematic tree in Fig. \ref{fig:landmarks}. Next, we explain networks details.

$\mathbf{J}$ and $\mathbf{L}$ are concatenated to form $\mathbf{JL}$ which has been estimated beforehand by DAE (i.e. $\mathbf{JL}_{DAE}$). For the easiness of the reading we omit the subscript $._{DAE}$ from $\mathbf{JL}_{DAE}$ in the next formulations. Given a kinematic tree $\kappa\in\mathbb{R}^{42}$, we define $\mathbf{N}$ as:
\vspace{-0.1cm}
\begin{equation}\label{eq:N}
    \mathbf{N}_i=\frac{\mathbf{JL}_{i} - \mathbf{JL}_{\kappa(i)}}{\| \mathbf{JL}_{i} - \mathbf{JL}_{\kappa(i)} \|_2}, \text{ for } i\in[2..42],\vspace{-0.1cm}
\end{equation}
where $\kappa(i)$ defines parenthood indices. The reason for this normalization is that, in order to compute relative joint rotation $\mathbf{R}$, we do not need to know relative distances. This frees network capacity from unnecessary data variability. Such relative distances are embedded in the computation of shape parameters. Thus, given relative distances $\mathbf{B}^*$ computed from template joints and landmarks $\mathbf{JL}^*$, we define $\mathbf{B}$ as:
\vspace{-0.1cm}
\begin{equation}
    \mathbf{B}_i^*=\| \mathbf{JL}_{i}^* - \mathbf{JL}_{\kappa(i)}^* \|_2 , \text{ for } i\in[2..42],
    \vspace{-0.1cm}
\end{equation}
\begin{equation}
    \mathbf{B}_i=\| \mathbf{JL}_{i} - \mathbf{JL}_{\kappa(i)} \|_2 - \mathbf{B}_i^*, \text{ for } i\in[2..42].
    \label{eq:bone}\vspace{-0.1cm}
\end{equation}
SMPL originally provides two different models for male and female. Selecting a proper gender model for each sample has a crucial impact on the accuracy at inference time. Therefore, we also include gender as an extra term on the shape parameters $\beta$. We assign gender a value from the set $\{-1,+1\}$ and learn it as a regression problem along with other shape parameters. 


\textbf{Loss function.} $\mathcal{L}_2$ loss has been commonly applied in recent regression problems \cite{tan2017indirect,pavlakos2018learning}. However, we found $\mathcal{L}_2$ loss to have problems in convergence and generalization in case of noisy inputs. Instead we use $\mathcal{L}_1$ loss on $\mathcal{R}$ and $\beta$, called $L_R$ and $L_\beta$, to supervise $\Omega$ and $\Psi$ networks. Firstly, this is done isolated from SMPL, which means no back-propagation is applied through SMPL. This is important for a stable training and fast convergence. Then, for performance gains, we fine-tune the networks adding $\mathcal{L}_1$ loss on SMPL output, called $L_{SMPL}$. SMPL output contains $\mathbf{J}$ and $\mathbf{T}$. However, to have SMPLR architecture resembling an autoencoder, we compute $L_{SMPL}$ on landmarks rather than the whole $\mathbf{T}$. The final SMPLR loss is $L_R+L_\beta+L_{SMPL}$.


\vspace{-0.1cm}
\subsection{Denoising autoencoder}
\vspace{-0.1cm}
Estimated joints by any CNN may have structured noise. For instance, in the case of occluded joints the error is higher due to their ambiguity and the lack of visual evidence. Visible joint predictions have as well structured error, following a Gaussian distribution. Such structured or Gaussian noise can be detected and learnt explicitly, helping to further improve initial estimation of $\mathbf{JL}_{CNN}$ to be fed into SMPLR module. Denoising autoencoder networks \cite{vincent2010stacked} are useful tools for such scenario, being able to learn structured patterns of the input better than ordinal autoencoders.

In this paper, we propose a DAE network as a bridge between CNN backbone and SMPLR. With the proposed DAE we are able to denoise 3D joints and landmarks. This procedure can be critical for error-prone CNNs, such as shallow networks. Moreover, it can be detached from CNN and trained independently given a large amount of mocap or synthetic SMPL-generated data. However, DAE may not generalize well to the noisy test data if it is trained with noise-free ground truth data. Therefore, it is important to train DAE with adversarial noise in this scenario for generalization purposes. In the section \ref{sec:dae} we show it is possible to train DAE with constrained uniform or Gaussian noise mimicking structured error without loss of generalization. It is also possible that only 2D joints are annotated in a given dataset. In sections \ref{sec:dae} and \ref{sec:sota} we also show DAE can lift 2D estimations to 3D. This could also be done by the use of mocap or synthetic 3D data projected to 2D and adversarial noise. The architecture of DAE is shown in Fig. \ref{fig:pipeline}. We apply two dropouts, one right after the input and the other after last encoder layer. 
By applying skip connections between encoder and decoder layers we force the network to learn noise structure in a fast and stable way. The input to DAE is the initial estimation of $\mathbf{JL}_{CNN}$ and the output is denoised $\mathbf{JL}_{DAE}$. We apply $\mathcal{L}_1$ loss (so called $L_{DAE}$) on $\mathbf{JL}_{DAE}$ to train this network. In order to force the awareness of adjacent joints correlations, we applied an $\mathcal{L}_1$ loss on the relative joints ($\mathbf{B}$ in Eq. \ref{eq:bone}) as well. However, we observed no significant impact on the results.

\begin{table*}[!t]\renewcommand{\arraystretch}{0.9}

\centering
\begin{tabular}{|l?p{1.7cm}|p{0.9cm}|p{0.9cm}|p{0.9cm}|p{0.9cm}|p{0.9cm}|p{0.9cm}|p{0.9cm}|p{0.9cm}|p{0.9cm}|p{0.8cm}|p{0.8cm}|}
  \hline
   & \multicolumn{1}{c|}{Model} & \multicolumn{1}{c|}{Hd} & \multicolumn{1}{c|}{Ts} & \multicolumn{1}{c|}{Sr} & \multicolumn{1}{c|}{Ew} & \multicolumn{1}{c|}{Wt} & \multicolumn{1}{c|}{Hp} & \multicolumn{1}{c|}{Kn} & \multicolumn{1}{c|}{Ft} & \multicolumn{1}{c|}{Avg.} & \multicolumn{1}{c|}{Avg.} & \multicolumn{1}{c|}{Avg.} \\ 
   &  &  &  &  &  &  &  &  &  & Jt & Lm & Bn \\\hline \hline
   \parbox[t]{2mm}{\multirow{5}{*}{\rotatebox[origin=c]{90}{CNN}}} 
   & ${Alexnet}$ & 100.0 & 41.6 & 99.0 & 179.9 & 246.9 & 34.6 & 138.5 & 217.3 & 133.0 & - & 31.9 \\ \cline{2-13} 
   & ${SHN}_{nL}$ & 47.5 & 23.0 & 44.2 & 77.2 & 112.0 & 16.3 & 61.9 & 102.7 & 62.8 & - & 10.4 \\ \cline{2-13}
   & $SHN$ & 46.2 & 23.0 & 43.0 & 75.5 & 110.2 & 15.3 & 61.2 & 102.2 & 59.9 & 61.5 & 9.3 \\ \cline{2-13}
   & ${SHN}_{e2e}$ & 45.1 & 22.2 & 43.3 & 74.4 & 108.2 & 16.0 & 57.9 & 94.2 & 57.8 & 59.6 & \textbf{9.0} \\ \cline{2-13}
   & ${SHN}^{final}$ & \textbf{40.8} & \textbf{20.9} & \textbf{38.0} & \textbf{66.8} & \textbf{93.4} & \textbf{14.3} & \textbf{55.7} & \textbf{92.9} & \textbf{53.0} & \textbf{54.3} & 9.7  
\\ \hline \hline
   \parbox[t]{2mm}{\multirow{3}{*}{\rotatebox[origin=c]{90}{DAE}}} 
   & ${DAE}_{Alexnet}$ & 89.3 & 37.1 & 87.2 & 160.8 & 230.8 & 29.6 & 131.7 & 205.9 & 121.5 & - & 22.7 \\ \cline{2-13}
   & ${DAE}_{SHN}$ & \textbf{45.8} & \textbf{22.2} & \textbf{42.2} & \textbf{75.4} & \textbf{108.5} & \textbf{14.4} & \textbf{61.8} & \textbf{103.2} & \textbf{59.2} & \textbf{61.1} & \textbf{9.5} \\ \cline{2-13}
   & ${DAE}_{SHN}^{2d}$ & 51.4 & 23.1 & 46.2 & 83.7 & 121.7 & 15.1 & 66.6 & 115.4 & 65.2 & 66.1 & 11.8 \\ \hline \hline
   \parbox[t]{2mm}{\multirow{3}{*}{\rotatebox[origin=c]{90}{SMPLR}}} 
   & $\Psi$ & 16.5 & 9.1 & 13.8 & 17.3 & 19.9 & 5.8 & 11.7 & 21.3 & 14.4 & 11.5 & 6.6 \\ \cline{2-13}
   & $\Omega$ & 55.1 & 22.3 & 48.8 & 85.8 & 127.1 & 13.4 & 68.6 & 122.3 & 67.8 & - & - \\ \cline{2-13}
   & $\Omega_{smpl}$ & \textbf{50.1} & \textbf{20.1} & \textbf{44.9} & \textbf{83.6} & \textbf{123.6} & \textbf{12.4} & \textbf{63.9} & \textbf{111.9} & \textbf{63.8} & - & - \\ \hline \hline
   \parbox[t]{2mm}{\multirow{2}{*}{\rotatebox[origin=c]{90}{ALL}}} 
   & $ALL$ & 57.9 & \textbf{24.0} & 52.8 & 92.7 & 140.4 & \textbf{15.8} & 67.8 & 115.0 & 70.8 & 73.8 & 8.4 \\ \cline{2-13}
   & ${ALL}_{Proc}$ & \textbf{53.7} & 26.1 & \textbf{49.7} & \textbf{86.2} & \textbf{129.9} & 21.6 & \textbf{67.0} & \textbf{109.2} & \textbf{67.8} & \textbf{70.6} & \textbf{7.7} \\ \cline{2-13}
   \hline
  
\end{tabular}\vspace{-0.2cm}
\caption{Ablation study of model components. Error in mm. Hd:Head, Ts:Torso, Sr:Shoulder, Ew:Elbow, Wt:Wrist, Hp:Hip, Kn:Knee, Ft:Foot, Jt:Joints, Lm:Landmarks and Bn:Bone length, $\{.\}_{nL}$: training without landmarks, $\{.\}^{final}$: training with limb heatmaps and data augmentation, $\{.\}_{Alexnet}$: $Alexnet$ estimations as input, $\{.\}_{SHN}$: $SHN$ estimations as input, $\{.\}^{2d}$: input depth is set to 0, $\{.\}_{smpl}$: training with $L_R+L_{SMPL}$ loss, $\{.\}_{e2e}$: model after end-to-end training, $\{.\}_{Proc}$: results after Procrustes mapping. Best results are bolded.\vspace{-0.4cm}
}
\label{tab:ablation}
\end{table*}

\vspace{-0.1cm}
\section{Experiments}
\label{sec:experiments}
\vspace{-0.1cm}
This section describes training details, datasets, and evaluation protocol. Then, we perform an ablation study of the different model components and compare it against state-of-the-art alternatives.
\vspace{-0.1cm}

\subsection{Training details}
\label{sec:setup}
\vspace{-0.1cm}

We build our backbone CNN based on the well-known stacked hourglass network (SHN) \cite{newell2016stacked} using 5 stacks. We extend final layers of each stack to volumetric heatmaps, i.e. including an extra dimension to discretize depth of the joints into 16 bins. The output of each stack is a tensor of size $64\times64\times16\times41$, where 64 is the size of X-Y axis and 41($=23+18$) is the number of joints and landmarks. We train this network with softmax cross entropy loss. All models and experiments were implemented on TensorFlow and trained on a GTX 1080 Ti. We used Adam optimizer in all the experiments with a learning rate of $0.01$ for SHN and $0.001$ for DAE, $\Omega$ and $\Psi$ networks. All networks are trained from scratch using a Xavier initializer. SHN converged in $150$-$250$ epochs with batch size $6$-$10$ samples. The rest of networks in the ablation analysis were trained with batch size $256$. We used a keeping probability 0.8 for dropout layer in DAE.

\textbf{Preprocessing.} Images are cropped to a square. To do so, we assume camera focal length and object distance to camera is available beforehand. First, the corners of a $2.5\times2.5$m grid, centered at average joint location and perpendicular to camera axis, are projected to image plane and define the cropping area. Then, cropped images are scaled to network input size ($256\times256$). This enforces a proportionality among pixel and real world sizes, larger people will appear bigger in image space as well. In those cases where the crops land outside the frame, a random image from VOC Pascal dataset is used for padding. Following \cite{varol18_bodynet} we use ground truth focal length and object distance to the camera in all experiments, both in training and inference time. However, to study the impact of scale ambiguity on the 3D joint prediction, we also estimate the cropping area in Human3.6M \cite{h36m_2014} dataset and show the results (see section \ref{sec:sota}).

\textbf{End-to-end training.} We applied incremental training. First, all networks (i.e. SHN, DAE, $\Omega$ and $\Psi$) were trained independently and then the whole network was fine-tuned end-to-end. In the ablation study we analyze the effect of different combinations of modules in the training.


\vspace{-0.1cm}
\subsection{Datasets}
\vspace{-0.1cm}

\textbf{UP-3D \cite{lassner2017unite}.} This dataset was designed by fitting a gender neutral SMPL model into images from LSP, LSP-extended and MPII-HumanPose datasets, keeping samples with better estimates. This yields a total of $8515$ labeled images in the wild, splitted into $5703$ for training, $1423$ for validation and $1389$ for test. Every sample is provided with 2D joints annotations and SMPL parameters.


\textbf{SURREAL \cite{varol17_surreal}.} Synthetic dataset of humans generated with SMPL model, containing exact annotations. 
It is composed of $68$K videos containing SMPL generated humans moving on top of random backgrounds. For sampling, we skip a frame if the average joint distance is lower than $5$cm w.r.t. to last sampled frame. This results in $2.8$M training, $27$K validation and $665$K test samples.


\textbf{Human3.6M \cite{h36m_2014}.} Human3.6M is a large dataset offering high precision 3D data thanks to MoCap sensors and calibrated cameras. It is composed of RGB videos of $11$ subjects performing $15$ actions twice while being recorded from $4$ different viewpoints. It contains around $3.6$ million frames. We sampled 1 of every 5 frames, ending with $312$K training and $110$K validation samples. Following state-of-the-art works, we use subjects S1, S5, S6, S7 and S8 for training, and S9 and S11 for testing. We generated ground truth SMPL parameters from the 3D data and body scans available in the dataset. Body scans allow an accurate estimation of shape parameters, computed only once per subject (shape does not change in short periods of time). Afterwards, we empirically defined a correspondence between SMPL joints and available 3D MoCap data. This matching is not perfect for some joints, which are weighted between $[0.25, 0.75]$ empirically to provide good estimations. This matching allows optimization of pose through an $\mathcal{L}_2$ loss. Finally, correspondence of back joints is not accurate, so instead, we use a loss that penalizes unrealistic back bends, by correcting back's pose parameters that land outside an empirically defined symmetrical range centered at 0.

\vspace{-0.1cm}
\subsection{Evaluation protocol}
\vspace{-0.1cm}
We evaluate the models by mean per joint position error (MPJPE) in millimeters (mm). The same metric is extended to surface points to report error of the generated body meshes. Following related works we apply two protocols: \textbf{Protocol 1} where all joints/points are subtracted from the root joint and, \textbf{Protocol 2} where estimated joints are aligned with ground truth through Procrustes analysis. We also report mean Intersection over Union (IoU) on body silhouette after mesh projection to the image plane.

\vspace{-0.1cm}
\subsection{Ablation study}
\vspace{-0.1cm}
In this section, we study different components of the proposed model on SURREAL validation set. For this task we subsample the training dataset into 89K frames such that every pair of samples has at least one joint displaced 150mm w.r.t. to each other, thus enforcing a uniform distribution over the whole dataset. We use the setup in Sec. \ref{sec:setup} to train each component. We explored several combination strategies during training to see the impact of each on the validation set. Except end-to-end training, all building blocks are trained isolated from the rest. We show results and description of each module in Tab. \ref{tab:ablation}.

\vspace{-0.1cm}
\subsubsection{CNN backbone}
\vspace{-0.1cm}
We first evaluate the performance of the CNN backbones. The results are shown in Tab. \ref{tab:ablation} under \textbf{CNN} row. We first train a baseline $Alexnet$ to regress 3D joints (without landmarks) using $L_2$ loss, Adam optimizer, learning rate 0.01 and batch size 32. We chose $Alexnet$ for two reasons: 1) to compare the results with the proposed volumetric SHN, and 2) it is a shallow network and prone to have structured error so that we can study DAE impact as well. As expected $Alexnet$ is not performing well to directly regress 3D joints. We then train volumetric SHN to predict $\mathbf{JL}$ and $\mathbf{J}$ (so-called ${SHN}$ and ${SHN}_{nL}$, respectively). As a result, landmarks help ${SHN}$ to gain 3mm improvement over ${SHN}_{nL}$. Next we explain our contributions to the default volumetric SHN.

\begin{figure}[!t]
    \centering
    \begin{subfigure}[b]{0.45\textwidth}
        \includegraphics[width=\textwidth]{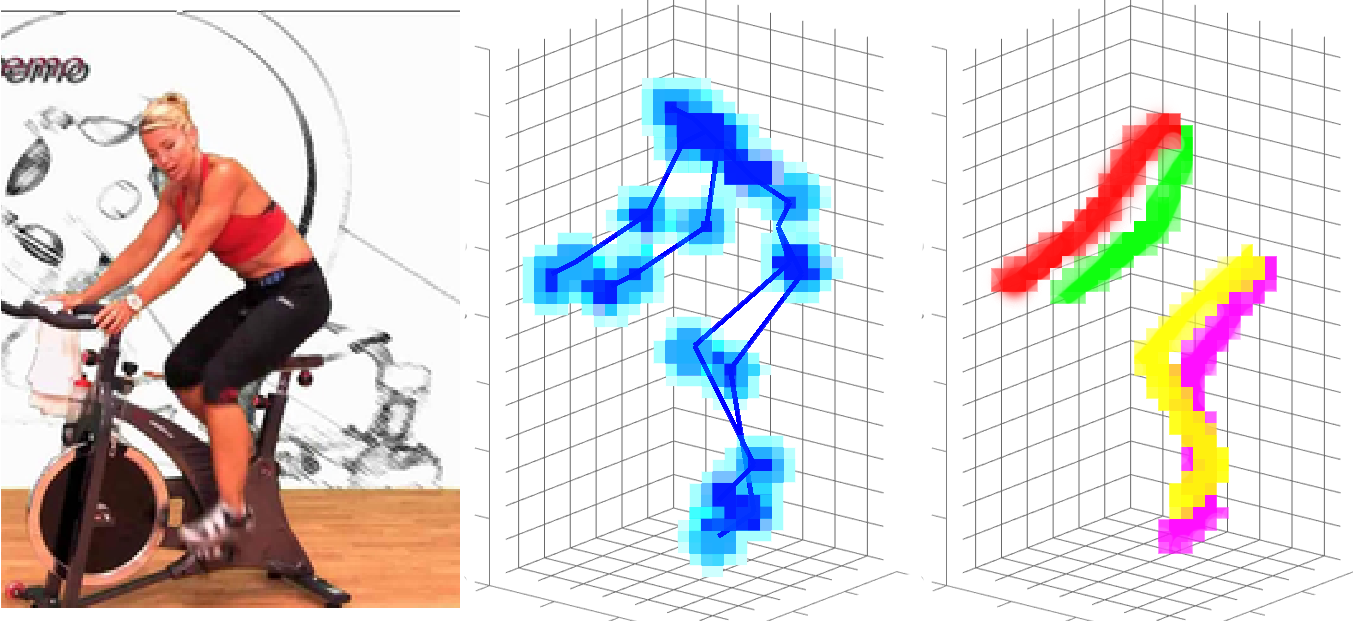}
    \end{subfigure}\vspace{-0.3cm}
    \caption{Sample volumetric heatmap of joints (middle) and limbs (right), each limb coded with a different color.}
    \label{fig:heatmap}
\end{figure}

\textbf{Final volumetric heatmap model.} To evaluate our method against state-of-the-art we extend default volumetric SHN to include limb heatmaps in the output and train the model using data augmentation. Limb heatmaps are $4$ additional volumetric heatmaps in the outputs of ${SHN}$. These heatmaps correspond to limb representations, created by composing segments from joint to joint (see Fig. \ref{fig:heatmap}). By fitting these heatmaps we expect to enforce the model to learn spatial relationships among joints to improve generalization. 

Besides regular data augmentation (including random color noise, flipping and rotation), two extra methods are applied: random background and artificial occlusion. By using binary masks for subjects provided at each frame, we remove the default background and replace it with a random image from VOC Pascal dataset. Similarly, we place random objects from VOC Pascal on random locations of the image to artificially create occlusions \cite{sarandi2018augmentation}. In both cases we do not use images containing humans. We call this model $SHN^{final}$. The results displayed in the Tab. \ref{tab:ablation} show how these simple strategies indeed enhance the performance of default ${SHN}$ by about 7mm on average joint error. 

\vspace{-0.1cm}
\subsubsection{Denoising autoencoder} \label{sec:dae}
\vspace{-0.1cm}
We also evaluate DAE trained with different inputs in several scenarios. Results are shown in Tab. \ref{tab:ablation} under \textbf{DAE} row.

\textbf{Could we train DAE independent to SHN?} Since DAE sequentially appears after SHN, it receives estimations from SHN. To answer this question we train DAE, as input, with i) 3D ground truth joints plus uniform noise with adapted bounds for each joint and ii) 3D joints estimated by ${SHN}$ (so-called ${DAE}_{noise}$ and ${DAE}_{SHN}$). We then evaluate both models with ${SHN}$ estimations as input at test time. As a result, ${DAE}_{noise}$ has an average error of 61.7mm (not shown in the table) which is similar to ${DAE}_{SHN}$ (61.9mm). This shows the generalization ability of DAE.

\textbf{Is DAE able to recover from structured noise?}  Other than ${DAE}_{SHN}$, we also train and test DAE with $Alexnet$ estimations (called ${DAE}_{Alexnet}$). For $Alexnet$ predicitons, DAE improves the error by 11mm, while on $SHN$ the improvement is 0.7mm. This shows the ability of DAE to learn structured error.

\textbf{Is DAE able to lift 2D joints to 3D?} To answer this question, we train and test DAE, following \cite{martinez2017baseline}, with ${SHN}$ estimations while depth is set to 0 (called ${DAE}_{SHN}^{2d}$). In fact, we want to test how DAE performs in the lack of 3D ground truth data. As a result, the average error is slightly higher than 65mm. Although the average error is 3mm higher than $SHN$, it shows DAE can lift 2D pose to 3D with successful results. We note that training ${DAE}_{SHN}^{2d}$ converges way slower than ${DAE}_{SHN}$.

\vspace{-0.1cm}
\subsubsection{SMPLR}
\vspace{-0.1cm}
In this section, we evaluate different components of SMPLR using ${SHN}$ estimations as input in both training and test. The results are shown in Tab. \ref{tab:ablation} under \textbf{SMPLR} row. We first evaluate the impact of shape and pose estimations isolated from each other within SMPLR. In test time, shape and pose estimations are fed into SMPL to evaluate final joints error.

\textbf{Shape estimation.} We train $\Psi$ network with $L_\beta$ loss. During test we feed estimated $\beta$ along with ground truth \textbf{$\mathcal{R}$} to SMPL. The results are shown as $\Psi$ in Tab. \ref{tab:ablation}. As one can see shape estimation has a low impact on final error (around 14mm avg. joints error).

\textbf{Pose estimation.} We train $\Omega$ network first with $L_R$ loss and then fine-tune it with $L_R+L_{SMPL}$ loss. The results are shown as $\Omega$ and $\Omega_{smpl}$ in Tab. \ref{tab:ablation}, respectively. During test we feed estimated $\mathcal{R}$ along with ground truth \textbf{$\beta$} to SMPL. As a result we gain 4mm improvement in pose estimation by applying $L_R+L_{SMPL}$ loss. In general, the higher source of errors in SMPLR is in pose parameters rather than shape.

\begin{figure*}[!ht]
    \centering
    \begin{subfigure}{.33\textwidth}
        \includegraphics[width=5.5cm,height=3.1cm]{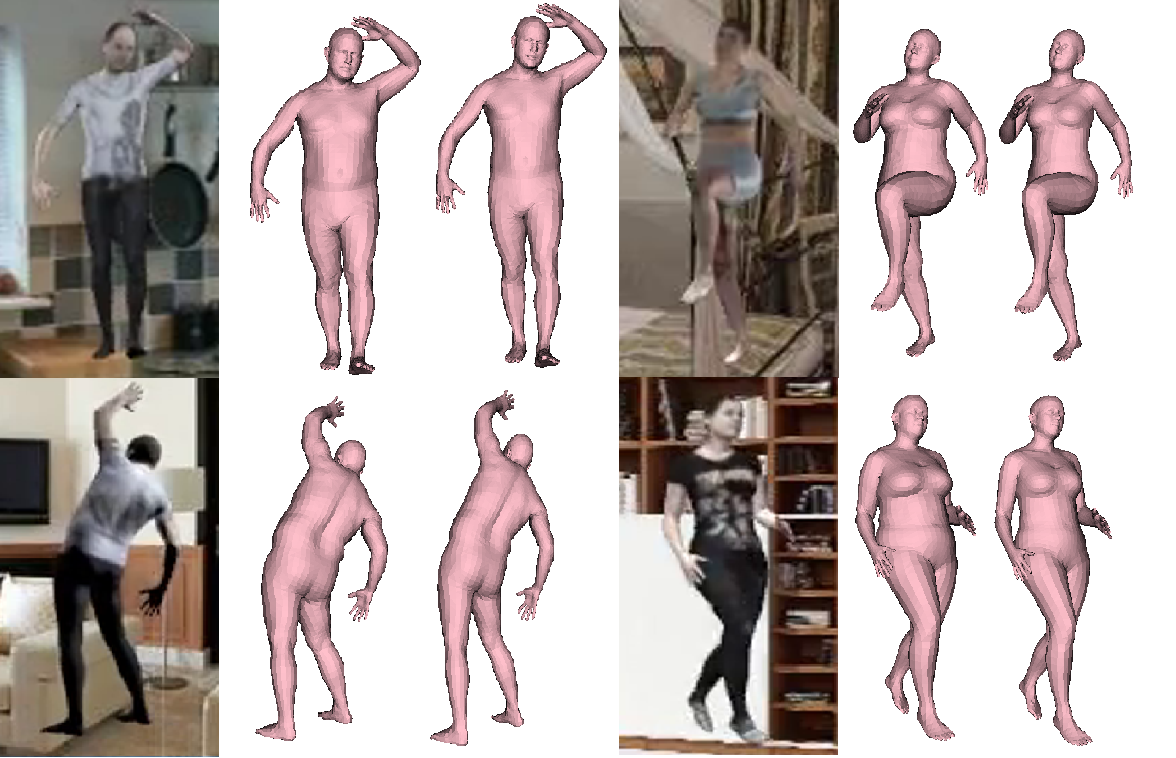}
        \label{fig:landmarks_shape}
        \caption{}\vspace{-0.4cm}
    \end{subfigure}
    \begin{subfigure}{.33\textwidth}
        \includegraphics[width=5.5cm,height=3.1cm]{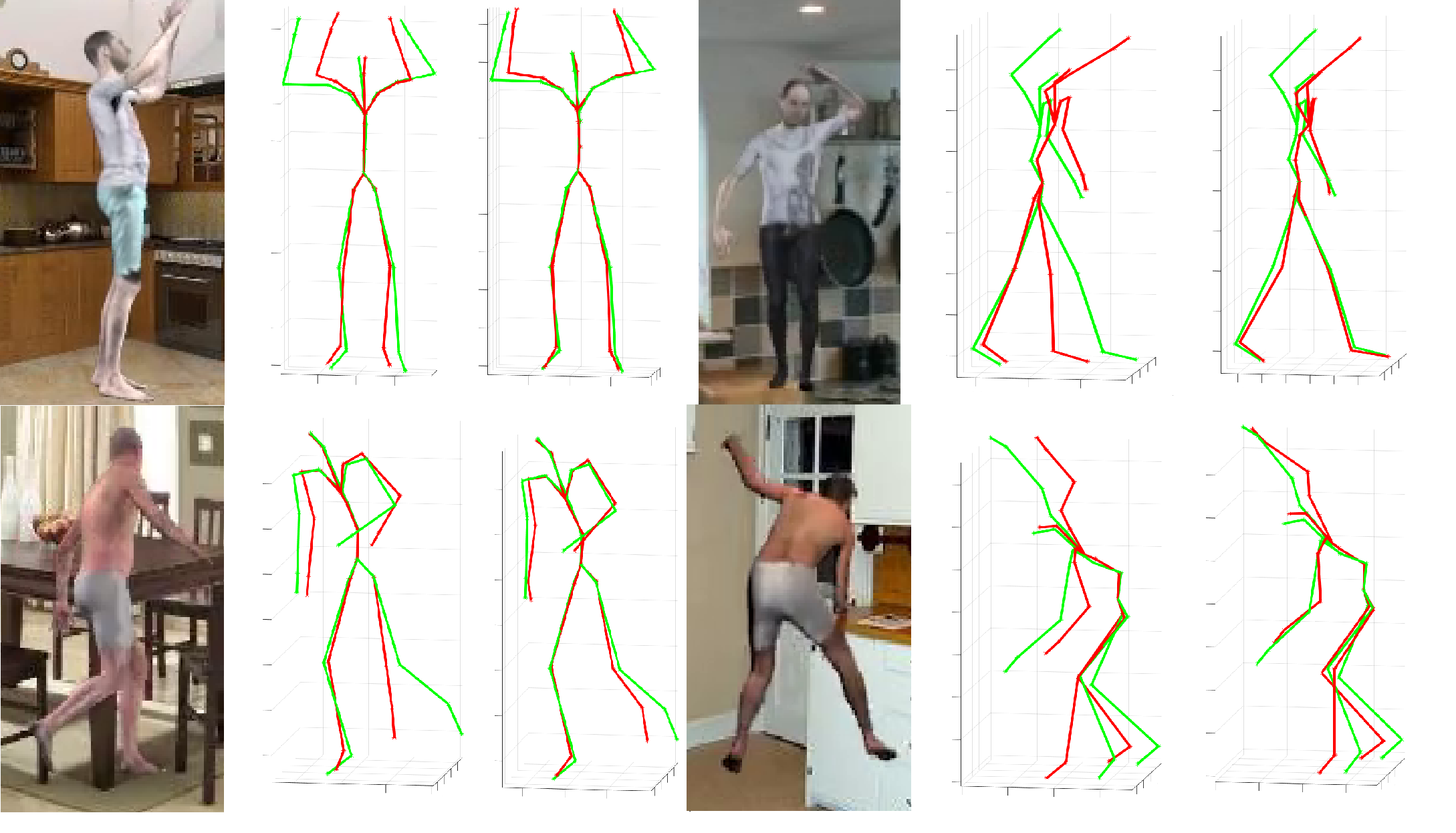}
        \label{fig:e2e_train}
        \caption{}\vspace{-0.4cm}
    \end{subfigure}
    \begin{subfigure}{.33\textwidth}
        \includegraphics[width=5.5cm,height=3.1cm]{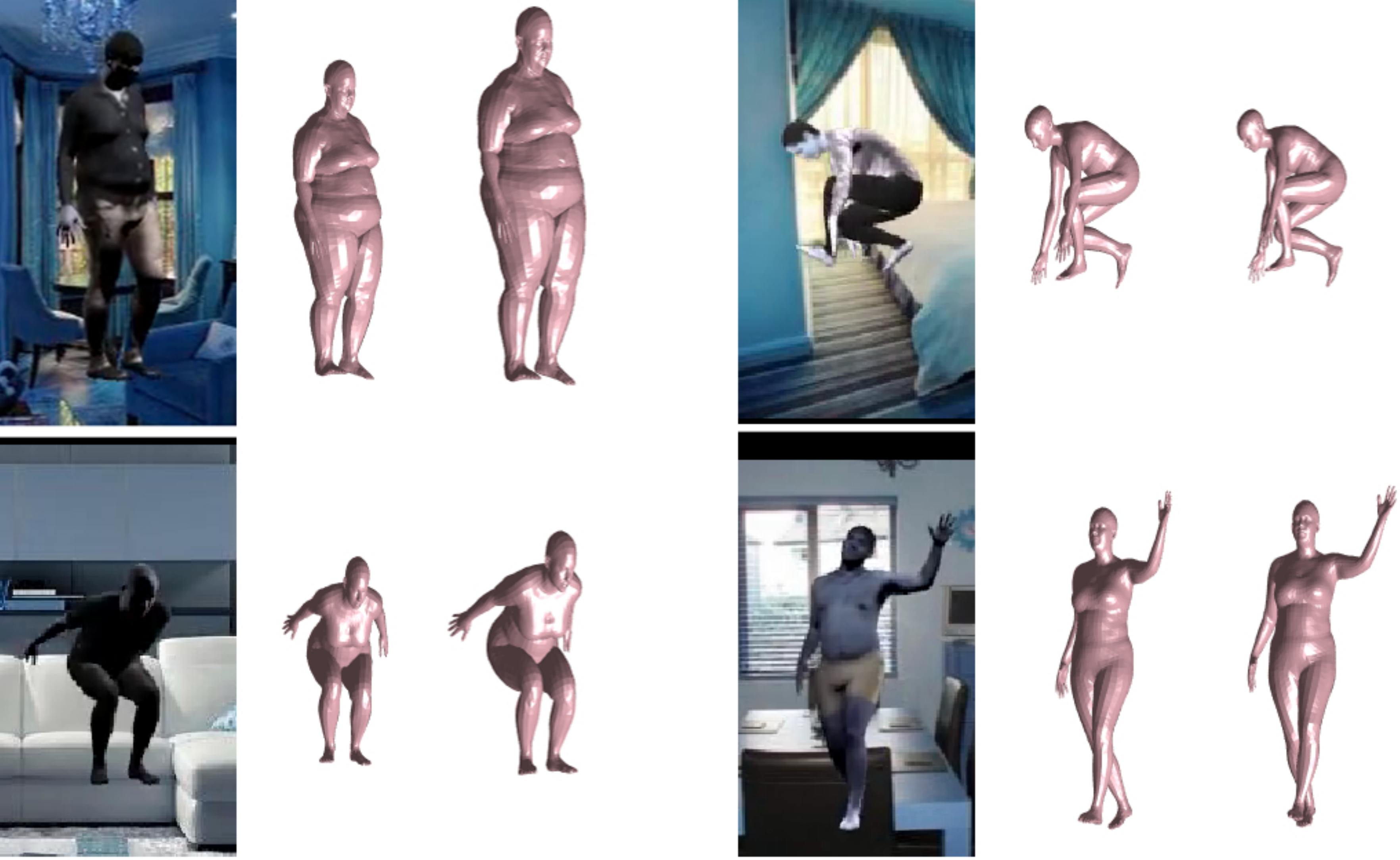}
        \label{fig:gender_procrustes}
        \caption{}\vspace{-0.4cm}
    \end{subfigure}
    \caption{Qualitative results of the ablation analysis. a) Visualization of the improvement on shape estimation due to landmarks. Left: image. Middle: estimation without landmarks. Right: estimation with landmarks. b) Prediction improvement due to end-to-end training. Left: image. Middle: prediction before end-to-end training. Right: prediction after end-to-end training. Green and red skeletons correspond to ground truth and predictions, respectively. c) Random samples with mistaken gender or viewpoint. Left: image. Middle: SMPL mesh. Right: SMPL mesh after Procrustes.}\label{fig:abc}\vspace{-0.1cm}
\end{figure*}

\textbf{Impact of landmarks.} Landmarks provide more visual evidence to the CNN when they are available in the dataset. Comparing $SHN$ to ${SHN}_{nL}$ in Tab. \ref{tab:ablation}, one can see landmarks improve head, arms and hip estimations. We also train $\Psi$ network with and without landmarks. Some qualitative results are shown in Fig.~\ref{fig:abc}(a).

\textbf{Gender.} We evaluate the accuracy of gender estimation in $\Psi$ and achieve 89.5\% accuracy. Such a high accuracy is critical for SMPL rendering. This means a given vector of shape parameters is interpreted differently by each gender model, i.e. a correctly estimated shape parameter but wrong gender estimation produces a wrong mesh generation, introducing a high error in SMPL mesh.

%
%

\setlength{\tabcolsep}{1pt} 
\begin{table}[!t]\centering
\begin{tabular}{l|c|c|l|c|c|}
\hline
\multicolumn{1}{|l|}{}              & \multicolumn{1}{l|}{Prot. 1} & \multicolumn{1}{l|}{Prot. 2} &           & \multicolumn{1}{l|}{Prot. 1} & \multicolumn{1}{l|}{Prot. 2} \\ \hline
\multicolumn{1}{|l|}{Bogo \cite{bogo2016keep}}          & \multicolumn{1}{c|}{-}       & 82.3                         & Tome \cite{tome2017lifting}      & 88.4                         & 70.7                         \\ \hline
\multicolumn{1}{|l|}{Lassner \cite{lassner2017unite}}       & \multicolumn{1}{c|}{-}       & 80.7                         & Pavlakos \cite{pavlakos2017coarse}  & 71.9                         & 51.9                         \\ \hline
\multicolumn{1}{|l|}{Tung \cite{tung2017self}${^*}$} & 98.4                         & \multicolumn{1}{c|}{-}       & Zhou \cite{zhou2017towards}     & 64.9                         & -                            \\ \hline
\multicolumn{1}{|l|}{Pavlakos \cite{pavlakos2018learning}} & \multicolumn{1}{c|}{-}       & 75.9                         & Martinez \cite{martinez2017baseline}  & 62.9                         & 47.7                   \\ \hline
\multicolumn{1}{|l|}{Omran \cite{omran2018neural}}         & \multicolumn{1}{c|}{-}       & 59.9                         & Sun \cite{sun2017compositional}      & 59.1                         & 48.3                          \\ \hline
\multicolumn{1}{|l|}{Kanazawa \cite{kanazawa2018end}}      & \multicolumn{1}{c|}{87.9}    & 56.8                         & Sun \cite{sun2018integral}${^{**}}$      & 64.1                         & -                         \\ \hline \hline
\multicolumn{6}{|l|}{Errors when cropping image is estimated} \\ \hline
\multicolumn{1}{|l|}{$ALL$}         & \multicolumn{1}{c|}{68.7}    & 52.5                         & $SHN^{final}$ & 62.2                         & 50.6                         \\ \hline
\multicolumn{1}{|l|}{${ALL}_{Proc}$}         & \multicolumn{1}{c|}{63.5}    & 52.5                         & $SHN_{e2e}^{final}$ & 57.4                     & 46.8          \\ \hline \hline
\multicolumn{6}{|l|}{Errors with ground truth cropping image} \\ \hline
\multicolumn{1}{|l|}{$ALL$}         & \multicolumn{1}{c|}{67.9}    & 52.2                         & $SHN^{final}$ & 61.3                         & 50.1                         \\ \hline
\multicolumn{1}{|l|}{${ALL}_{Proc}$}         & \multicolumn{1}{c|}{\textbf{62.6}}    & \textbf{52.2}                         & $SHN_{e2e}^{final}$ & \textbf{56.5}                     & \textbf{46.3}          \\ \hline
\end{tabular}\vspace{-0.2cm}
\caption{MPJPE error in mm. on Human3.6M for both protocols. Best results are bolded. Left columns: SMPL-based methods comparison. Right columns: state-of-the-art 3D pose comparison. SMPLR outperforms all SMPL-based methods, and the simple proposed SHN updates show state-of-the-art-results after end-to-end training without using any extra data. SMPL surface errors on this dataset are 88.2 and 81.3mm for $ALL$ and $ALL_{Proc}$, respectively. * Tung \etal \cite{tung2017self} use 32 joints. ** Sun \etal \cite{sun2018integral} report the results with and without extra data in the training. For a fair comparison we take this number from where no extra data has been used.}\vspace{-0.1cm}
\label{tab:sota-h36m}
\end{table}

\begin{table}[!t]\centering
\begin{tabular}{|l|c|c|}
\hline
 & SMPL surface & 3D joints \\ \hline
Tung \cite{tung2017self} & 74.5 & 64.4 \\ \hline
Varol (independent) \cite{varol18_bodynet} & 74.5 & 46.1$^*$ \\ \hline
Varol (multi-task) \cite{varol18_bodynet} & 65.8 & \textbf{40.8}$^{*}$ \\ \hline \hline
$SHN^{final}$ & - & 42.8$^*$ \\ \hline
$SHN_{e2e}^{final}$ & - & \textbf{40.8}$^{*}$ \\ \hline
${ALL}$ & 66.0 & 50.6 \\ \hline
${ALL}_{Proc}$ & \textbf{62.3} & 48.2 \\ \hline
\end{tabular}\vspace{-0.2cm}
\caption{Errors (mm) on SURREAL dataset (protocol 1). Best results are bolded. $.^{*}$ are intermediate estimated 3D joints used to predict SMPL surface.}
\label{tab:sota-surreal}
\end{table}

\begin{table}[!t]\centering
\begin{tabular}{|l|c|c|c|}
\hline
 & SURREAL & Human3.6M & UP-3D \\ \hline
Varol (multi-task) \cite{varol18_bodynet} & - & - & 0.73 \\ \hline
${ALL}_{Proc}$ & \textbf{0.75} & \textbf{0.71} & \textbf{0.77} \\ \hline
\end{tabular}\vspace{-0.2cm}
\caption{Silhouette IoU on three datasets.}
\label{tab:iou}
\end{table}

\begin{table}[h!]
\centering
\begin{tabular}{|l|l|l|l|l|}
\hline
                & 2D SHN   & Vol. SHN   & DAE   & SMPLR \\ \hline
Train             & 6 / 2.8 & 6 / 16.8 & 256 / 0.054 & 256 / 2.5      \\ \hline
Test             & 6 / 0.35 & 6 / 3.1 & 256 / 0.011 & 256 / 0.87      \\ \hline
Test             & 1 / 0.32 & 1 / 0.36 & 1 / 0.007 & 1 / 0.32      \\ \hline
\end{tabular}
\vspace{-0.3cm}\caption{
Processing time (batch size/time in sec.) of 1 step with 41 heatmaps/points. 2D SHN is the default stack hourglass network for 2D joints estimation. Both 2D and vol. SHN have 5 stacks. Vol. SHN has 16 depth bins. SMPLR includes SMPL on top.}
\label{tab:time}\vspace{-0.3cm}
\end{table}






\begin{figure*}[ht!]
    \centering
    \includegraphics[width=\textwidth]{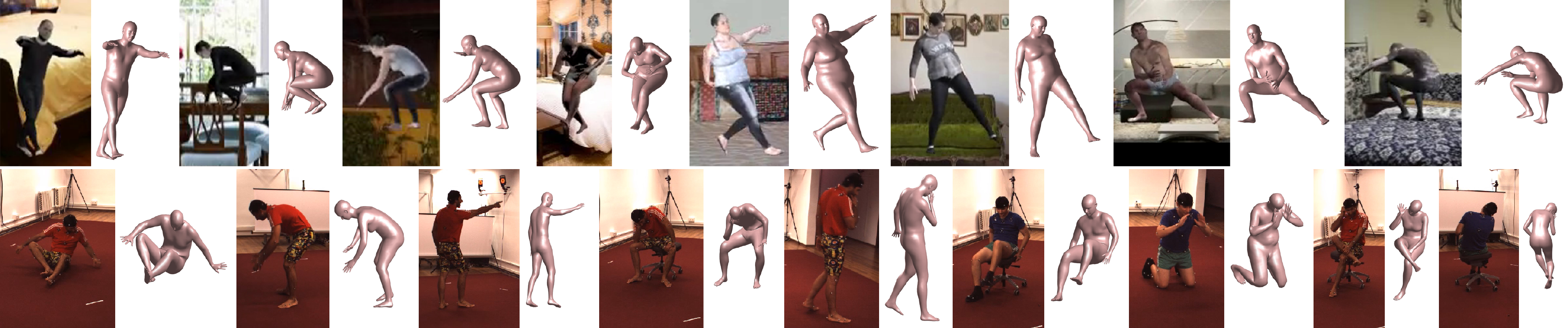}
\vspace{-0.6cm}    \caption{Qualitative results. Top: SURREAL. Bottom: Human3.6M. Last sample of each row shows a failure case due to inaccurate pose estimation produced mainly by occlusion and/or background confusion.}
    \label{fig:qualitative1}\vspace{-0.1cm}
\end{figure*}

\begin{figure*}[ht!]
    \centering
    \includegraphics[width=17.3cm,height=3.5cm]{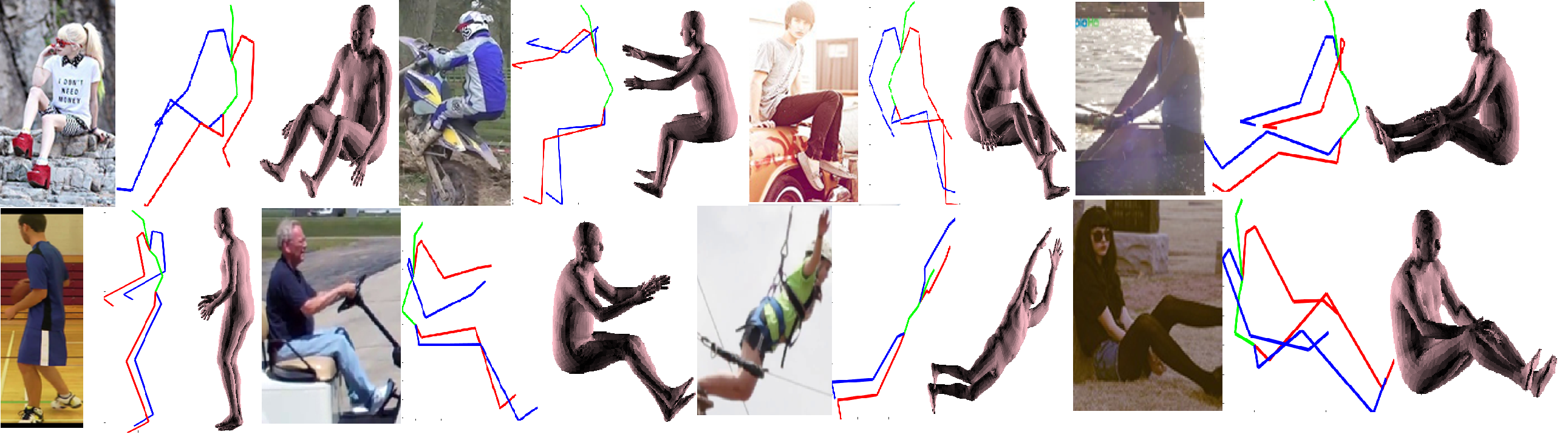}\vspace{-0.3cm}
    \caption{UP-3D results. Based on estimated 2D joints (middle) we compute 3D joints and render mesh (right).
    }\vspace{-0.4cm}
    \label{fig:qualitative2}
\end{figure*}

\vspace{-0.1cm}
\subsection{End-to-end training}\label{sec:end-to-end ablation}
\vspace{-0.1cm}
Here, we describe how the end-to-end training was performed. Thanks to soft-argmax, the model is differentiable and trainable end-to-end. We first explore a regular end-to-end training by stacking already trained models $SHN$, $DAE_{SHN}$, $\Psi$ and $\Omega_{smpl}$ along with SMPL on top. The loss is a summation of all intermediate losses. The order of magnitude of SHN loss is several times lower than the other losses. Therefore, without a proper balancing, the weights of the ${SHN}$ vanish after few training steps. We empirically set this balance to be around $1e$-$5$. Fine-tuning is performed with a low learning rate, empirically set to $1e$-$4$, to ensure learning stability. We observed this model does not show improvements. Therefore, we propose the next procedure for end-to-end training.

We first train $DAE$, $\Psi$ and $\Omega_{smpl}$ with ground truth 3D joints until they overfit on training data. Once trained, they are frozen and appended to ${SHN}$. Fine-tuning is performed with the same loss balancing and learning rate as before. 
The network shows improvement after the first training epoch, and after an additional epoch it fully converges. To ensure that this improvement is the result of the proposed end-to-end training, we trained ${SHN}$ alone for more than $10$ additional epochs without showing any improvement. The results of $SHN$ after end-to-end training in Tab. \ref{tab:ablation} (called ${SHN}_{e2e}$) shows more than 2mm improvement. Some qualitative results are shown in Fig.~\ref{fig:abc}(b).

\textbf{Recover SMPLR error.} We stack all trained models to report final SMPL predictions (row \textbf{ALL} in Tab. \ref{tab:ablation}). SMPLR naturally has a generalization error. Wrongly estimated gender is a source of error in SMPLR. Not only gender, but also global rotation error embedded in $\Omega$ can degrade the results. Fortunately, the mesh can be partially corrected by an affine transformation as a post-processing. To do so we apply Procrustes mapping from SMPLR output to its input $\mathbf{JL}$ and update the mesh accordingly. The results in Tab. \ref{tab:ablation} shows 4mm error recovery of $ALL_{Proc}$ vs. $ALL$. Some qualitative examples are shown in Fig.~\ref{fig:abc}(c). Note that for this post-processing step, no ground truth information is required, as we align SMPLR output with SHN output (both model predictions), knowing that SHN is more accurate. This is different from protocol 2, where final predictions are aligned with ground truth. 

\vspace{-0.1cm}
\subsection{State-of-the-art comparison}\label{sec:sota}
\vspace{-0.1cm}
\textbf{Human3.6M.} We compare our results to state-of-the-art on Human3.6M in Tab. \ref{tab:sota-h36m}, split in two sets: SMPL-based solutions vs. only 3D pose recovery. In the former, we outperform state-of-the-art for both evaluation protocols, especially in protocol 1 improving \cite{kanazawa2018end} over 25mm. We note that we use $\Psi$ network trained on SURREAL dataset to estimate shape parameters, since Human3.6M contains just 5 subjects in the training set (therefore, only 5 shapes). The results in the second set show that our simple modifications to SHN achieve state-of-the-art results after end-to-end training (${SHN}_{e2e}^{final}$). Compared to \cite{pavlakos2017coarse}, a fixed small depth resolution of 16 bins in the volumetric heatmap works better than a coarse-to-fine setup. As we mentioned earlier, we also want to study the results when image cropping is not based on ground truth data. Therefore, we estimate camera focal length and object distance to camera following \cite{sarandi2018augmentation} and use them to compute the cropping. The error on cropped image is less than 5px on each corner. To be robust against this error we fine-tune ${SHN}^{final}$ with an additional random scaling image augmentation. As a result, the 3D joints error is increased less than 1mm in average which is quite marginal on this dataset. Fig. \ref{fig:qualitative1} shows some qualitative results.

\textbf{SURREAL.} The competitors on this dataset are \cite{varol18_bodynet} and \cite{tung2017self}. Note that \cite{varol18_bodynet} relies on all ground truth data in a multi-task setup to generate volumetric body shape which limits applicability. That is, volumetric body shape is a coarse representation and not parametric. The results in Tab. \ref{tab:sota-surreal} show we outperform \cite{varol18_bodynet} by 3.5mm for SMPL surface error. Interestingly, our ${SHN}_{e2e}^{final}$ achieves same 3D joints estimation error ($40.8$ mm) than \cite{varol18_bodynet} without performing multi-task learning. 
We show some qualitative results in Fig. \ref{fig:qualitative1}.

\textbf{UP-3D.} We use this dataset to show results in a in-the-wild scenario lifting 2D joints to 3D. To do so, we fine-tune ${SHN}^{final}_{e2e}$ pre-trained on SURREAL. Note that we do not train this model end-to-end on UP-3D dataset. Since this dataset is of small size, we include a subset of 18K images from SURREAL for training. During training the input to DAE is ground truth joints plus uniform noise and depth is set to 0. During testing, SHN estimations (with depth set to 0) are fed to DAE. The inputs to SMPLR are DAE estimations. The SMPL errors in test set before and after Procrustes mapping are around 91mm and 87mm, respectively, being below the 100.5mm error reported in \cite{pavlakos2018learning}. 
Fig. \ref{fig:qualitative2} shows some qualitative results.

\textbf{Silhouette IoU.} To check the quality of rendered body, we also compute silhouette IoU of the estimated mesh after projection to image plane. The results can be seen in Tab. \ref{tab:iou}. We achieve a high IoU (more than 0.7) on all datasets without explicitly training the network for this task.

\vspace{-0.1cmF}
\subsection{Time complexity}
\vspace{-0.2cm}

We show the processing time of each module of the proposed network in Tab. \ref{tab:time}. Experiments were conducted on a GTX1080TI. We compare a default 2D SHN with our proposed volumetric SHN (both SHNs have 5 stacks). As expected, volumetric SHN is 6 times slower than 2D SHN in the training for a batch size of 6. However, there is not much difference between them at inference time for a batch size of 1. We must note that 2D SHN can be fit in GPU memory for a batch size of 12 while volumetric SHN can have at most a batch size of 6. Our SMPLR implementation can be run in 3 FPS for a batch size of 1 at inference time.

\vspace{-0.1cm}
\section{Conclusions}
\label{conclu}
\vspace{-0.2cm}

We proposed a deep-based framework to recover 3D pose and shape from a still RGB image. Our model is composed of a SHN backbone followed by a DAE and a network capable of reversing SMPL from sparse data. Such model, capable of end-to-end training, is able to accurately reconstruct human body mesh. We experimentally found that processing SHN output joints with DAE removes structured error. We have also shown that SMPL model can be reversed and used to recover 3D pose and shape. 
Finally, we exploit SMPLR capabilities in the training of deep learning networks by backpropagating SMPL related errors through the SHN. We evaluated our proposal on SURREAL and Human3.6M datasets and improved SMPL-based state-of-the-art alternatives by 3.5 and 25 mm, respectively on each dataset.

\section*{Acknowledgements}

This work has been partially supported by the Spanish project TIN2016-74946-P (MINECO/FEDER, UE) and CERCA Programme / Generalitat de Catalunya. We gratefully acknowledge the support of NVIDIA Corporation with the donation of the GPU used for this research. This work is partially supported by ICREA under the ICREA Academia programme.

{\small
\bibliographystyle{ieee}
\bibliography{main}

\begin{thebibliography}{10}\itemsep=-1pt

\bibitem{anguelov2005scape}
D.~Anguelov, P.~Srinivasan, D.~Koller, S.~Thrun, J.~Rodgers, and J.~Davis.
\newblock Scape: shape completion and animation of people.
\newblock In {\em ACM TOG}, volume~24, pages 408--416, 2005.

\bibitem{bogo2016keep}
F.~Bogo, A.~Kanazawa, C.~Lassner, P.~Gehler, J.~Romero, and M.~J. Black.
\newblock Keep it smpl: Automatic estimation of 3d human pose and shape from a
  single image.
\newblock In {\em ECCV}, pages 561--578, 2016.

\bibitem{brau20163d}
E.~Brau and H.~Jiang.
\newblock 3d human pose estimation via deep learning from 2d annotations.
\newblock In {\em 3D Vision (3DV)}, pages 582--591. IEEE, 2016.

\bibitem{chen20173d}
C.-H. Chen and D.~Ramanan.
\newblock 3d human pose estimation= 2d pose estimation+ matching.
\newblock In {\em CVPR}, pages 5759--5767. IEEE, 2017.

\bibitem{guan2009estimating}
P.~Guan, A.~Weiss, A.~O. Balan, and M.~J. Black.
\newblock Estimating human shape and pose from a single image.
\newblock In {\em ICCV}, pages 1381--1388. IEEE, 2009.

\bibitem{hoang2018cascade}
V.-T. Hoang and K.-H. Jo.
\newblock 3d human pose estimation using cascade of multiple neural networks.
\newblock In {\em IEEE Transaction on Industrial Informatics}, 2018.

\bibitem{h36m_2014}
C.~Ionescu, D.~Papava, V.~Olaru, and C.~Sminchisescu.
\newblock Human3.6m: Large scale datasets and predictive methods for 3d human
  sensing in natural environments.
\newblock {\em IEEE TPAMI}, 36(7):1325--1339, jul 2014.

\bibitem{ivekovivc2008human}
{\v{S}}.~Ivekovi{\v{c}}, E.~Trucco, and Y.~R. Petillot.
\newblock Human body pose estimation with particle swarm optimisation.
\newblock {\em Evolutionary Computation}, 16(4):509--528, 2008.

\bibitem{kanazawa2018end}
A.~Kanazawa, M.~J. Black, D.~W. Jacobs, and J.~Malik.
\newblock End-to-end recovery of human shape and pose.
\newblock In {\em CVPR}, 2018.

\bibitem{lassner2017unite}
C.~Lassner, J.~Romero, M.~Kiefel, F.~Bogo, M.~J. Black, and P.~V. Gehler.
\newblock Unite the people: Closing the loop between 3d and 2d human
  representations.
\newblock In {\em IEEE CVPR}, pages 4704--4713, 2017.

\bibitem{loper2015smpl}
M.~Loper, N.~Mahmood, J.~Romero, G.~Pons-Moll, and M.~J. Black.
\newblock Smpl: A skinned multi-person linear model.
\newblock {\em ACM Transactions on Graphics (TOG)}, 34(6):248, 2015.

\bibitem{luo2018orinet}
C.~Luo, X.~Chu, and A.~Yuille.
\newblock Orinet: A fully convolutional network for 3d human pose estimation.
\newblock {\em BMVC}, 2018.

\bibitem{martinez2017baseline}
J.~Martinez, R.~Hossain, J.~Romero, and J.~J. Little.
\newblock A simple yet effective baseline for 3d human pose estimation.
\newblock In {\em ICCV}, 2017.

\bibitem{mehta2017vnect}
D.~Mehta, S.~Sridhar, O.~Sotnychenko, H.~Rhodin, M.~Shafiei, H.-P. Seidel,
  W.~Xu, D.~Casas, and C.~Theobalt.
\newblock Vnect: Real-time 3d human pose estimation with a single rgb camera.
\newblock {\em ACM Transactions on Graphics (TOG)}, 36(4):44, 2017.

\bibitem{moreno20173d}
F.~Moreno-Noguer.
\newblock 3d human pose estimation from a single image via distance matrix
  regression.
\newblock In {\em CVPR}, pages 1561--1570. IEEE, 2017.

\bibitem{newell2016stacked}
A.~Newell, K.~Yang, and J.~Deng.
\newblock Stacked hourglass networks for human pose estimation.
\newblock In {\em ECCV}, pages 483--499, 2016.

\bibitem{nibali20183d}
A.~Nibali, Z.~He, S.~Morgan, and L.~Prendergast.
\newblock 3d human pose estimation with 2d marginal heatmaps.
\newblock {\em arXiv preprint arXiv:1806.01484}, 2018.

\bibitem{omran2018neural}
M.~Omran, C.~Lassner, G.~Pons-Moll, P.~Gehler, and B.~Schiele.
\newblock Neural body fitting: Unifying deep learning and model based human
  pose and shape estimation.
\newblock In {\em 3D Vision (3DV)}, pages 484--494. IEEE, 2018.

\bibitem{pavlakos2017coarse}
G.~Pavlakos, X.~Zhou, K.~G. Derpanis, and K.~Daniilidis.
\newblock Coarse-to-fine volumetric prediction for single-image 3d human pose.
\newblock In {\em CVPR}, pages 1263--1272. IEEE, 2017.

\bibitem{pavlakos2018learning}
G.~Pavlakos, L.~Zhu, X.~Zhou, and K.~Daniilidis.
\newblock Learning to estimate 3d human pose and shape from a single color
  image.
\newblock {\em CVPR}, 2018.

\bibitem{sarandi2018augmentation}
I.~Sa\'ra\'ndi, T.~Linder, K.~O. Arras, and B.~Leibe.
\newblock Synthetic occlusion augmentation with volumetric heatmaps for the
  2018 eccv posetrack challenge on 3d human pose estimation.
\newblock In {\em ECCV PoseTrack Workshop}, 2018.

\bibitem{sigal2008combined}
L.~Sigal, A.~Balan, and M.~J. Black.
\newblock Combined discriminative and generative articulated pose and non-rigid
  shape estimation.
\newblock In {\em Advances in neural information processing systems}, pages
  1337--1344, 2008.

\bibitem{sigal2010humaneva}
L.~Sigal, A.~O. Balan, and M.~J. Black.
\newblock Humaneva: Synchronized video and motion capture dataset and baseline
  algorithm for evaluation of articulated human motion.
\newblock {\em International journal of computer vision}, 87(1-2):4, 2010.

\bibitem{sigal2012loose}
L.~Sigal, M.~Isard, H.~Haussecker, and M.~J. Black.
\newblock Loose-limbed people: Estimating 3d human pose and motion using
  non-parametric belief propagation.
\newblock {\em International journal of computer vision}, 98(1):15--48, 2012.

\bibitem{stoll2011fast}
C.~Stoll, N.~Hasler, J.~Gall, H.-P. Seidel, and C.~Theobalt.
\newblock Fast articulated motion tracking using a sums of gaussians body
  model.
\newblock In {\em ICCV}, pages 951--958. IEEE, 2011.

\bibitem{sun2017compositional}
X.~Sun, J.~Shang, S.~Liang, and Y.~Wei.
\newblock Compositional human pose regression.
\newblock 2017.

\bibitem{sun2018integral}
X.~Sun, B.~Xiao, F.~Wei, S.~Liang, and Y.~Wei.
\newblock Integral human pose regression.
\newblock In {\em Proceedings of the European Conference on Computer Vision
  (ECCV)}, pages 529--545, 2018.

\bibitem{tan2017indirect}
V.~Tan, I.~Budvytis, and R.~Cipolla.
\newblock Indirect deep structured learning for 3d human body shape and pose
  prediction.
\newblock {\em BMVC}, 2017.

\bibitem{tome2017lifting}
D.~Tome, C.~Russell, and L.~Agapito.
\newblock Lifting from the deep: Convolutional 3d pose estimation from a single
  image.
\newblock {\em CVPR 2017 Proceedings}, pages 2500--2509, 2017.

\bibitem{tung2017self}
H.-Y. Tung, H.-W. Tung, E.~Yumer, and K.~Fragkiadaki.
\newblock Self-supervised learning of motion capture.
\newblock In {\em Advances in Neural Information Processing Systems}, pages
  5236--5246, 2017.

\bibitem{varol18_bodynet}
G.~Varol, D.~Ceylan, B.~Russell, J.~Yang, E.~Yumer, I.~Laptev, and C.~Schmid.
\newblock {BodyNet}: Volumetric inference of {3D} human body shapes.
\newblock In {\em ECCV}, 2018.

\bibitem{varol17_surreal}
G.~Varol, J.~Romero, X.~Martin, N.~Mahmood, M.~J. Black, I.~Laptev, and
  C.~Schmid.
\newblock Learning from synthetic humans.
\newblock In {\em CVPR}, 2017.

\bibitem{vincent2010stacked}
P.~Vincent, H.~Larochelle, I.~Lajoie, Y.~Bengio, and P.-A. Manzagol.
\newblock Stacked denoising autoencoders: Learning useful representations in a
  deep network with a local denoising criterion.
\newblock {\em Journal of machine learning research}, 11(Dec):3371--3408, 2010.

\bibitem{yang20183d}
W.~Yang, W.~Ouyang, X.~Wang, J.~Ren, H.~Li, and X.~Wang.
\newblock 3d human pose estimation in the wild by adversarial learning.
\newblock In {\em CVPR}, 2018.

\bibitem{zanfir2018deep}
A.~Zanfir, E.~Marinoiu, M.~Zanfir, A.-I. Popa, and C.~Sminchisescu.
\newblock Deep network for the integrated 3d sensing of multiple people in
  natural images.
\newblock In {\em Advances in Neural Information Processing Systems}, pages
  8410--8419, 2018.

\bibitem{zhou2017towards}
X.~Zhou, Q.~Huang, X.~Sun, X.~Xue, and Y.~Wei.
\newblock Towards 3d human pose estimation in the wild: a weakly-supervised
  approach.
\newblock {\em ICCV}, 2017.

\end{thebibliography}
}

\end{document}